\documentclass{ws-procs11x85}
\usepackage{ws-procs-thm}   

\newif\ifshowrebuttalcolor

\showrebuttalcolorfalse

\usepackage{tabularx}
\usepackage{xcolor}
\usepackage{hyperref}
\usepackage[breakable]{tcolorbox}
\usepackage{tabularx}
\usepackage{array}
\usepackage{xcolor}
\usepackage{booktabs}

\usepackage{adjustbox}
\usepackage{multirow}

\definecolor{rebuttalblue}{RGB}{0, 102, 204}

\newcommand{\rebuttal}[1]{%
  \ifshowrebuttalcolor
    \textcolor{rebuttalblue}{#1}%
  \else
    #1%
  \fi
}

\begin{document}

\title{Text Knows What, Tables Know When: Clinical Timeline Reconstruction via Retrieval-Augmented Multimodal Alignment}

\author{Sayantan Kumar$^{1\ast}$, Shahriar Noroozizadeh$^{2,3\ast}$, Juyong Kim$^{2\ast}$, Jeremy C. Weiss$^{1}$}

\address{
$^{1}$National Library of Medicine, National Institutes of Health, Bethesda, MD, USA\\
$^{2}$Machine Learning Department, Kenneth C. Griffin School of Computer Science, \\Carnegie Mellon University, Pittsburgh, PA, USA\\
$^{3}$Heinz College of Information Systems Management and Public Policy, \\Carnegie Mellon University, Pittsburgh, PA, USA\\
}

\begin{abstract}
Clinical language models increasingly operate over electronic health records (EHRs), yet patient records are not stored as temporally grounded trajectories. Clinical notes describe symptoms, assessments, and disease progression, but often compress or narratively reorder events. Structured EHR rows provide timestamps for labs, medications, vitals, and procedures, but capture only part of the clinical story. We formulate clinical timeline reconstruction as retrieval-augmented temporal grounding: constructing a patient trajectory by using narrative text for event semantics and structured rows as partial temporal evidence. We introduce a scaffolded workflow that extracts central narrative events, builds an initial temporal scaffold, attaches non-central events, and calibrates timestamps using retrieved structured EHR rows. We evaluate on 40 discharge summaries, including 15 i2b2-derived and 25 MIMIC-IV summaries, each with manual gold-standard timelines and aligned structured EHR data. 
\rebuttal{Across models, multimodal calibration left event match rates largely unchanged and generally improved temporal performance: mean paired case-level multimodal--unimodal differences were positive in 7 of 12 model-metric comparisons across concordance and AULTC, with none negative. However, uncertainty was substantial given the 40-case sample; paired case-level bootstrap intervals excluded zero only for the DeepSeek V3.2 AULTC improvement.}
A gap analysis shows that 35.1\% of text-derived events have no structured counterpart. These findings support treating structured EHR data as partial temporal evidence for narrative-derived patient trajectories.
\end{abstract}

\keywords{Clinical timeline reconstruction; electronic health records; clinical narratives; retrieval-augmented generation; multimodal alignment; large language models}

\copyrightinfo{$^{\ast}$ These authors contributed equally to this work.\\
Accepted for oral presentation at the Pacific Symposium on Biocomputing (PSB) 2027.\vspace{1mm}\\
\copyright\ 2026 The Authors. Open Access chapter published by World Scientific Publishing Company and distributed under the terms of the Creative Commons Attribution Non-Commercial (CC BY-NC) 4.0 License.}

\begin{figure}[t]
    \centering
    \includegraphics[width=0.85\textwidth]{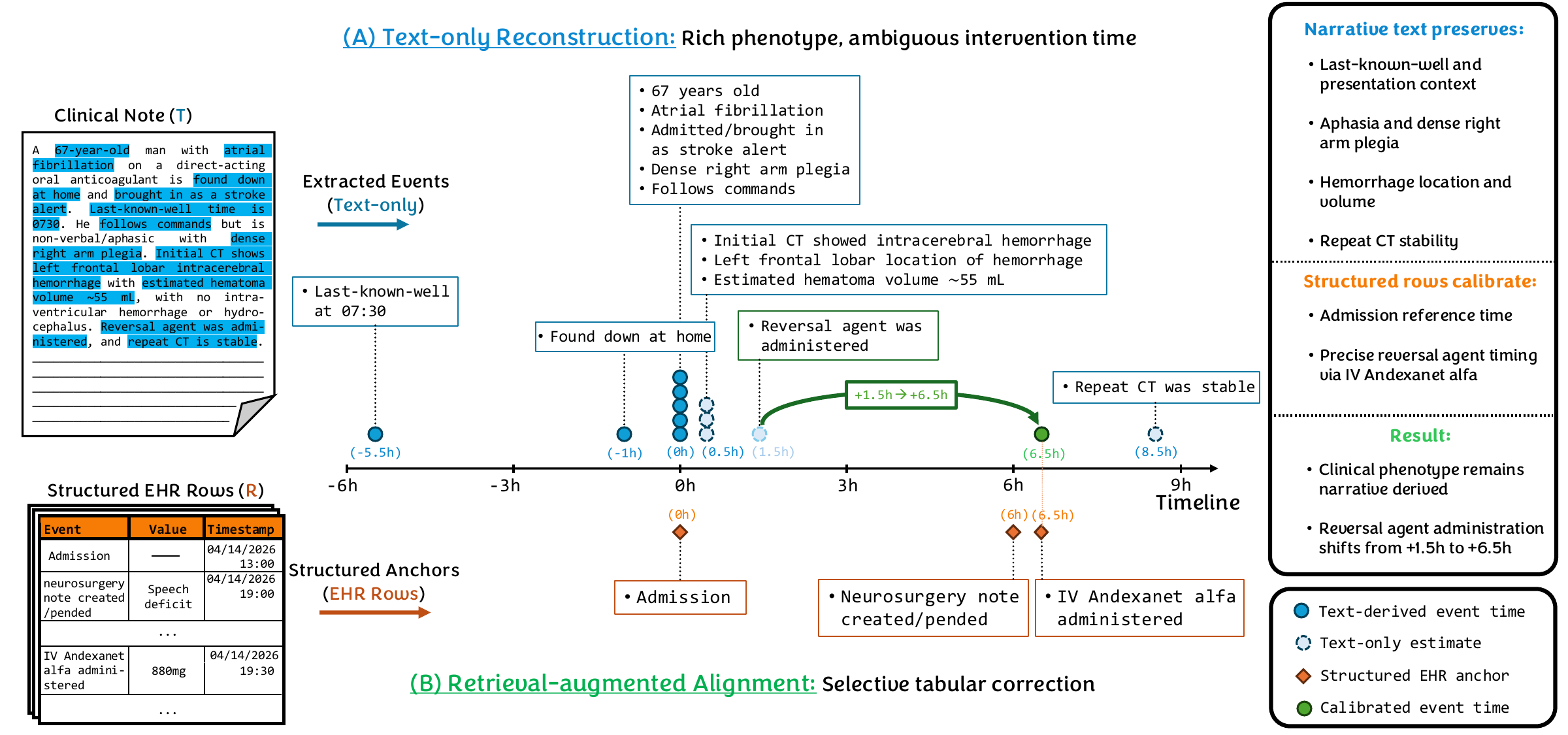}
    \vspace{-9pt}
    \caption{
    \rebuttal{Motivation for retrieval-augmented temporal grounding using a synthetic stroke-alert case constructed for illustration.}
    (\textbf{A}) Text-only reconstruction can recover clinically meaningful narrative events, but intermediate event times may remain ambiguous.
    (\textbf{B}) Retrieved structured EHR rows provide partial temporal evidence that can calibrate text-derived events when clinically relevant, improving temporal placement while preserving narrative event content.
    }
    \label{fig:motivation_multimodal_alignment}
    \vspace{-2mm}
\end{figure}

\vspace{-0.75em}
\section{Introduction}
\vspace{-0.25em}

Time is central to clinical reasoning in critical care.
Patients often follow rapidly evolving trajectories, requiring symptoms, physiologic measurements, treatments, and outcomes to be interpreted relative to admission, deterioration, intervention, and recovery. 
Sepsis trajectories exemplify this challenge.
Although Sepsis-3 remains important for trial eligibility and computational phenotyping \cite{kyriazopoulou2021procalcitonin,seymour2019derivation}, 
sepsis is better understood as a set of heterogeneous clinical trajectories with a shared dysregulatory response rather than as a single, uniform course. 
Similar temporal challenges extend across critical illness: risk forecasts depend on when evidence becomes available; treatment-response analyses require distinguishing pre- and post-intervention measurements; and retrospective cohort definitions may change when the timing of symptom onset, diagnostic testing, or treatment initiation is misaligned.
This increasing focus on temporal structure is evident in early-warning systems \cite{henry2022factors}, studies showing that predictive utility varies with timing \cite{kamran2024evaluation}, and research on the temporal characteristics of sepsis subtypes and progression \cite{noroozizadeh2023temporal}.

The growing emphasis on time in critical care reveals an important methodological gap: electronic health records (EHRs) are not organized as temporally grounded patient trajectories. Critical-care repositories such as MIMIC-III/IV \cite{johnson2016mimic,johnson2023mimic} make this challenge explicit by combining structured time-series measurements with narrative clinical documentation. These modalities provide complementary but asymmetric evidence for timeline construction (Figure~\ref{fig:motivation_multimodal_alignment}). Structured rows offer relatively precise temporal information from laboratory results, medications, procedures, vital signs, diagnoses, and administrative events. Clinical narratives, by contrast, provide richer accounts of the patient course, including symptoms, clinician assessments, patient-reported histories, functional status, disease progression, and contextual modifiers. For example, a note describing ``increasing lethargy and peripheral mottling'' may precede an increase in Sequential Organ Failure Assessment (SOFA) scores. Similarly, documented oliguria despite stable creatinine may indicate clinically meaningful states that structured measurements do not fully capture. 

Recent studies point to both the promise and the difficulty of this problem. Multimodal annotation of discharge summaries aligned with structured EHR data demonstrated that access to both modalities can improve the precision of inpatient event timelines, highlighting the value of structured rows as temporal evidence when aligned with the narrative record \cite{frattallone2024using}. At the same time, text-only timeline reconstruction remains challenging in realistic EHR-style notes, where dense event structure, abbreviations, and semi-structured formatting make both event recovery and time assignment difficult \cite{noroozizadeh2026reconstructing, noroozizadeh2026forecasting}. Together, these findings motivate more explicit alignment between text-based and tabular approaches, with the goal of improving temporal precision while preserving the richer event content available in narrative text.\cite{moldwin2021empirical,seinen2025using}

In this work, we investigate whether aligning retrieved structured EHR evidence with text-derived event timelines can improve the temporal precision of reconstructed patient trajectories. Our retrieval-augmented multimodal approach integrates structured rows directly into timeline reconstruction rather than treating textual and tabular data as separate views. The multistep pipeline first identifies temporally informative central events and uses them to create an initial scaffold of the patient trajectory. It then positions additional events relative to this scaffold and refines the resulting timeline using retrieved structured evidence. This decomposition reflects a clinically natural distinction between events that form the backbone of an encounter and those whose timing is best interpreted relative to that backbone. Using this framework, we assess whether and where multimodal alignment improves on text-only timeline reconstruction. Specifically, we examine the effects of introducing structured evidence at different stages of the reconstruction workflow, determine whether improvements are concentrated in temporal metrics rather than event recovery, and analyze which prognostically relevant information appears in text but is missing or delayed in tabular data.

\begin{figure}[t]
    \centering
    \includegraphics[width=\textwidth]{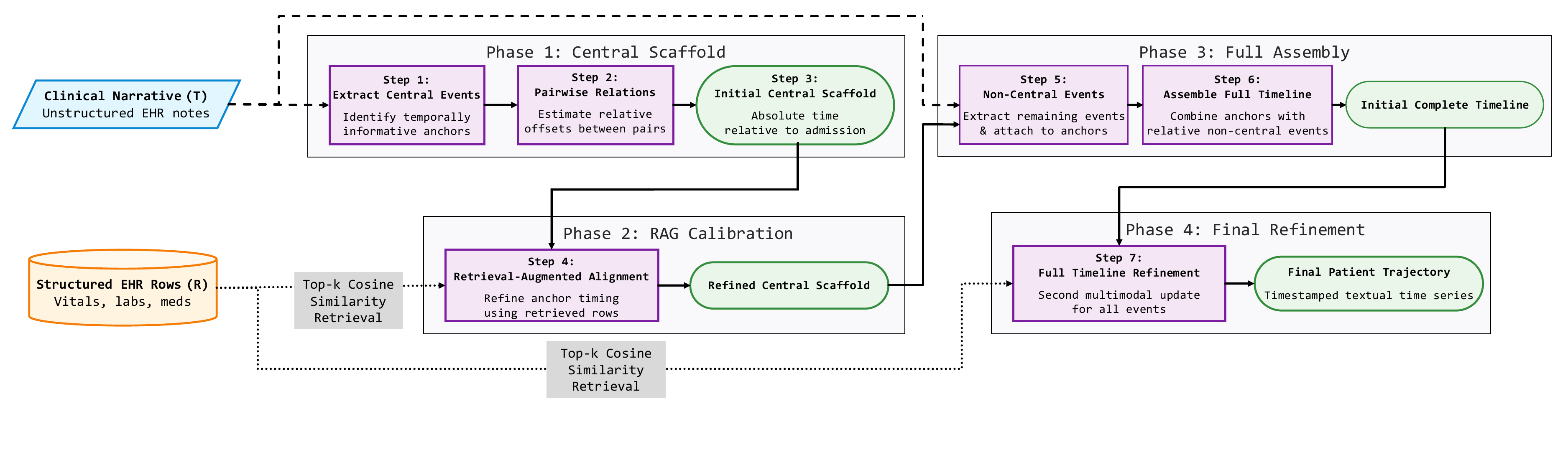}
    \vspace{-40pt}
    \caption{
    Overview of the scaffolded retrieval-augmented timeline reconstruction workflow. The method first extracts central events from the clinical narrative, estimates pairwise temporal relations, and builds an initial central-event scaffold. Retrieved structured EHR rows are then used to calibrate this scaffold. Non-central events are extracted and attached relative to central events, after which a final retrieval-augmented refinement step calibrates the complete timeline.
    }
    \label{fig:pipeline}
    \vspace{-10pt}
\end{figure}

\textbf{Contributions.} In summary, the paper makes the following contributions. \textbf{(i) Temporal grounding formulation:} We formulate clinical timeline reconstruction as retrieval-augmented temporal grounding of narrative-derived patient trajectories using structured EHR evidence. \textbf{(ii) Multimodal scaffolded reconstruction workflow:} We introduce a workflow that uses anchor events as a temporal backbone and retrieved structured rows as partial evidence for timestamp calibration. \textbf{ (iii) Multimodal timeline evaluation:} We evaluate multimodal calibration on 40 discharge summaries with aligned structured EHR rows and manual gold-standard timelines, testing whether structured evidence improves temporal localization more than event recovery. \textbf{(iv) Information missing from tabular data:} We provide an empirical analysis of clinically relevant events that are absent, delayed, or semantically compressed in structured tabular data, motivating why unstructured text is necessary for constructing clinically meaningful patient trajectories.

\vspace{-0.75em}
\section{Related Work}
\label{sec:related_work}
\vspace{-0.25em}

\textbf{Clinical temporal information extraction.}
Clinical timeline reconstruction builds on a long tradition of temporal reasoning and temporal information extraction in medicine \cite{kohane1987temporal,zhou2007temporal}. Early annotation schemes and shared tasks, including i2b2 and THYME, focused on identifying clinical events, time expressions, and temporal relations such as before, after, overlap, and containment \cite{uzuner20112010,sun2013evaluating,sun2013annotating,styler2014temporal}. Systems such as MedTime and cTAKES-temporal operationalized these ideas for clinical notes by extracting time expressions and temporal relations \cite{lin2013medtime,savova2010mayo,lin2016multilayered}. Pre-LLM work also studied patient timelines and longitudinal record summarization, including timeline generation from clinical notes and HARVEST for summarizing longitudinal patient records \cite{nikfarjam2013towards,hirsch2015harvest}. These studies established the importance of temporal structure in clinical narratives, but primarily focused on temporal annotation, relation extraction, or summarization rather than grounding narrative-derived absolute timelines in structured EHR evidence.

\vspace{5pt}
\noindent
\textbf{LLM-based clinical timeline reconstruction.}
Recent work has moved from relation-centered temporal extraction toward absolute event timelines. Leeuwenberg and Moens studied absolute event timeline extraction from English clinical reports, while Frattallone-Llado et al. showed that aligning discharge summaries with structured EHR data can improve inpatient event timeline precision \cite{leeuwenberg2020towards,frattallone2024using}. Textual time-series work has further shown that clinical narratives can be converted into event--time tuples for temporal forecasting, but these approaches are primarily text-only and do not use structured EHR rows as calibration evidence during reconstruction \cite{wang2025large,noroozizadeh2026reconstructing,noroozizadeh2026forecasting, kumar2026temporally}. Recent shared tasks such as ChemoTimelines have renewed interest in patient-level clinical timeline extraction using prompting, fine-tuning, and dictionary-enhanced LLM systems \cite{yao2025chemotimelines,zhang2025uwbionlp}. Our work differs from these efforts by treating timeline reconstruction as multimodal temporal grounding: narrative text supplies event semantics, while structured EHR rows are retrieved as partial temporal evidence for timestamp calibration.

\vspace{5pt}
\noindent
\textbf{Asymmetric evidence in structured and narrative EHR data.}
Clinical machine learning studies have shown that structured and unstructured EHR data encode complementary information for phenotyping, prediction, and patient representation \cite{moldwin2021empirical,liu2022multimodal,seinen2025using}. Structured data provide dense timestamped measurements, but often omit symptoms, functional status, patient-reported histories, and clinical reasoning. Narrative notes preserve this context, but are often retrospective and not organized as chronological trajectories. This asymmetry makes clinical timeline reconstruction different from standard multimodal fusion: the goal is not to merge two complete views of the same event sequence, but to use structured rows as partial temporal evidence for a narrative-derived patient trajectory. Our retrieval-augmented scaffolded workflow is designed for this setting, selecting candidate structured rows for each text-derived event and updating timestamps only when the retrieved evidence is clinically relevant.

\vspace{-0.75em}
\section{Methods}
\label{sec:methods}
\vspace{-0.25em}

\subsection{Clinical timeline task and dataset}
\label{sec:data}
\vspace{-0.25em}

Given a clinical narrative $T$ and associated structured EHR rows $R$, the goal is to reconstruct a textual time series $\hat{S}=\{(e_i,\hat{t}_i)\}_{i=1}^{N}$, where each $e_i$ is a standalone clinical event and each $\hat{t}_i \in \mathbb{R}$ is its predicted timestamp in hours relative to a case-specific reference time. When hospital admission is explicitly described, we use admission as $t=0$; otherwise, we use the earliest documented clinical encounter or presentation in the narrative. Events before the reference time receive negative timestamps, and events after it receive positive timestamps.

A clinical event is any patient-specific health-related mention that is semantically interpretable on its own and temporally locatable, including symptoms and signs, diagnoses, tests, procedures, treatments, medication administrations, major clinical states, outcomes, pertinent negatives, and termination or discontinuation events. Our event representation follows prior textual time-series work \cite{wang2025large,noroozizadeh2026reconstructing,noroozizadeh2025pmoa,kumar2026temporally}, but differs from short-span, concept-centric schemes such as i2b2 \cite{uzuner20112010,sun2013evaluating}. We represent events as standalone clinical findings: spans may include context needed for clinical specificity, conjunctive mentions are split when doing so improves clarity, and modifiers such as negation, uncertainty, severity, laterality, and intent are preserved when they affect event interpretation or temporal placement.

Our evaluation uses \textbf{40 discharge summaries} with manual gold-standard timelines and aligned structured EHR rows: \textbf{15 i2b2-derived discharge summaries} and \textbf{25 MIMIC-IV discharge summaries}. All cases have structured EHR rows aligned to the same hospital encounter. MIMIC-style critical-care records are well suited for this setting because they pair narrative clinical documentation with timestamped structured data, including laboratory values, medication administrations, vital signs, procedures, diagnoses, and administrative events \cite{johnson2016mimic,johnson2023mimic}. For each case, the narrative note provides the source text for event reconstruction, while the structured record provides external temporal evidence. We convert structured-row timestamps to hours relative to the same case-specific reference time used for the narrative timeline. Each structured row $r_m \in R$ is represented as $r_m=(\nu_m,x_m,s_m)$, where $\nu_m$ is the row name or clinical feature, $x_m$ is the observed value when available, and $s_m$ is the timestamp in hours relative to the reference time.

\rebuttal{The 15 i2b2-derived cases were randomly sampled from eligible cases; the 25 MIMIC-IV cases were sampled at the patient level, using the first discharge summary for five randomly sampled patients and one randomly selected summary for 20 others, without enrichment for diagnosis, temporal richness, or encounter complexity. Three medical experts annotated disjoint subsets; every case was annotated once from the note and aligned structured record, so no adjudication or inter-annotator agreement was available (\hyperref[supp:eval-cohort]{Supplementary Material \S\ref*{supp:eval-cohort}}).}

We treat structured rows as temporal evidence rather than gold-standard timeline events. Structured EHR tables contain dense repeated measurements, medication records, and machine-generated entries, many of which are not standalone narrative events. Conversely, clinically meaningful narrative events, such as symptoms, functional status, patient-reported histories, and clinical reasoning, may not have direct structured counterparts. Our method therefore uses structured rows to calibrate text-derived event times when clinically relevant, rather than forcing all structured rows into the reconstructed timeline.

\vspace{-0.75em}
\subsection{Scaffolded retrieval-augmented multimodal timeline reconstruction}
\label{sec:pipeline}
\vspace{-0.25em}

Our workflow is designed around three principles. First, it anchors before expanding: the model identifies temporally informative anchor events before reconstructing the full event set. Second, it retrieves candidate structured evidence rather than forcing every narrative event to align to a structured row. Third, \rebuttal{it uses preserve-by-default calibration}: event times are updated only when retrieved structured evidence is clinically relevant and temporally plausible. The workflow proceeds in seven clinically interpretable steps (Figure \ref{fig:pipeline}).\footnote{Step-specific prompts and the full reproducibility pipeline are available in our GitHub repository:\\\url{https://github.com/SayantanKumar/multimodal_textual_tabular_alignment}}
\vspace{-5pt}
\begin{enumerate}
    \item \textbf{Step 1: Extract anchor events.} The model first identifies temporally informative anchor events, also referred to as central events, that define the encounter backbone or serve as temporal referents for other events. Examples include admission, symptom onset, treatment initiation, clinical deterioration, procedures, transfers, and discharge. 

    \item \textbf{Step 2: Estimate timing among anchors.} The model estimates pairwise time differences (in hours) between selected anchor events.

    \item \textbf{Step 3: Build an initial anchor timeline.} The pairwise relations between anchor events are resolved into a provisional timeline backbone in hours relative to the case reference time. When estimates conflict, the model uses higher-confidence relations and the source narrative to construct the initial scaffold.

    \item \textbf{Step 4: Calibrate anchors using structured evidence.} For each anchor event, retrieved structured EHR rows are provided to the calibration model together with the narrative and provisional scaffold. The model updates an anchor time only when the structured row clinically supports the change; otherwise, the text-derived estimate is retained.

    \item \textbf{Step 5: Extract non-anchor supporting events relative to anchors.} After anchor calibration, the model extracts remaining non-anchor events and attaches each event to a referent anchor with a relative time offset and confidence score. These events include symptoms, exam findings, pertinent negatives, severity descriptors, functional status, medication details, and contextual clinical information.
    
    \item \textbf{Step 6: Assemble the full timeline.} The calibrated anchor times and relative supporting-event offsets are combined into an initial complete timeline, converting all events into absolute times in hours relative to the case reference time.
    
    \item \textbf{Step 7: Refine the complete timeline using structured evidence.} Structured rows are retrieved again for all reconstructed events, not only anchors. The model performs a final conservative calibration pass, updating timestamps when retrieved rows provide clinically consistent temporal evidence and preserving text-derived times otherwise. The output is a temporally grounded textual time series containing clinically meaningful narrative events with calibrated timestamps when structured evidence is relevant.
\end{enumerate}

\vspace{-10pt}
\paragraph{Retrieval and calibration module.}
The same retrieval module is used in Step 4 and Step 7. Structured rows are represented by concatenating the row event name and value; timestamps are excluded from the embedding text and passed separately to the calibration model. Rows with the same event name and value are merged, with timestamps represented in hours relative to the case reference time. We embed event strings and structured-row representations using MedTE (\texttt{mohammadkhodadad/MedTE-cl15-step-8000}), mean-pool and L2-normalize the embeddings, and rank rows by cosine similarity. For each event, we retrieve the top $k=5$ structured rows with no fixed similarity threshold. The calibration model receives each row as \texttt{name | value | time | similarity}. \rebuttal{Calibration follows a preserve-by-default rule: cosine similarity ranks candidates, irrelevant rows are disregarded, the initial time is retained when no relevant evidence justifies a change, and narrative context resolves repeated timestamps or conflicting rows. The complete decision rule appears in \hyperref[supp:calibration-rule]{Supplementary Material \S\ref*{supp:calibration-rule}}.}
We also evaluated $k=3$, $k=5$, and $k=10$, finding negligible performance differences but increased runtime for larger $k$.

This design treats retrieval as evidence selection rather than forced temporal alignment. Methods such as dynamic time warping assume that two sequences share an approximately comparable event universe and can be aligned through a monotone path. Clinical narratives and structured EHR rows often violate this assumption: many narrative events have no structured counterpart, while structured tables contain dense repeated measurements, medication records, and charted observations that are not standalone narrative events. Our workflow therefore allows structured evidence to calibrate event times when appropriate without requiring every narrative event to have a tabular match.

\vspace{-0.75em}
\subsection{Experimental setup}
\label{sec:experimental_setup}
\vspace{-0.25em}

\textbf{Baselines and ablations.} We compare the default scaffolded multimodal workflow against text-only, single-step, stage-specific, and forced-alignment baselines. \textbf{Text-only reconstruction} uses the same scaffolded workflow but removes both structured-EHR calibration steps, testing whether retrieved structured evidence improves temporal localization beyond narrative-only reconstruction. \textbf{Single-step multimodal reconstruction} directly generates a complete timeline from the narrative and retrieved structured rows without constructing an anchor-event scaffold, testing whether the staged anchor/non-anchor decomposition is useful beyond simply adding retrieved evidence to the prompt. To isolate where structured evidence contributes, we compare two stage-specific ablations: \textbf{anchor-only calibration}, which calibrates the anchor scaffold but omits final full-timeline calibration, and \textbf{final-only calibration}, which reconstructs the full timeline from text before applying structured-EHR calibration. The default workflow applies both anchor-stage and final-stage calibration. Finally, \textbf{dynamic time warping (DTW)} serves as a forced temporal-alignment baseline that attempts to align text-derived events with structured rows through sequence alignment rather than retrieval-based evidence selection followed by clinical calibration.

\vspace{5pt}
\noindent\textbf{Evaluation metrics.} We evaluate predicted timelines along two complementary dimensions: \emph{event recovery} and \emph{temporal localization}, adapted from prior textual time-series work \cite{wang2025large,noroozizadeh2026reconstructing}. Because predicted and reference timelines may differ in length and event wording, we first align predicted events to manually annotated reference events using a recursive best-match procedure. This produces a one-to-one alignment between predicted and reference events while accommodating unequal timeline lengths. \textbf{Event match rate} is the proportion of reference clinical events successfully aligned to a predicted event. Temporal metrics are computed on the matched subset. \textbf{Temporal concordance} (c-index) measures whether the predicted timeline preserves the relative ordering of events in the manual reference. \textbf{Area Under the Log-Time Cumulative Distribution Function (AULTC)} summarizes absolute timestamp discrepancy on a log-transformed time scale, with higher values indicating smaller timestamp errors across clinically meaningful ranges. Because concordance and AULTC are computed only over matched events, they should be interpreted jointly with event match rate. 
We estimate uncertainty using case-level bootstrap confidence intervals. Each bootstrap sample resamples discharge summaries with replacement and recomputes event matching and timeline metrics, preserving within-case dependence among events. We report 95\% confidence intervals for the reconstruction metrics using 200 bootstrap resamples.
\rebuttal{For direct multimodal--unimodal comparisons, we additionally used 10,000 paired case-level resamples and report
confidence intervals for the mean within-case differences in \hyperref[supp:paired-bootstrap]{Supplementary Material \S\ref*{supp:paired-bootstrap}}, \hyperref[tab:supp-paired-bootstrap]{Table~\ref*{tab:supp-paired-bootstrap}}.} 
\rebuttal{All implementation and reproducibility details are presented in \hyperref[supp:implementation]{Supplementary Material \S\ref*{supp:implementation}}.}

\vspace{-0.75em}
\section{Results}
\label{sec:results}
\vspace{-0.25em}

\subsection{\rebuttal{Structured EHR evidence calibrates event timing while preserving event recovery}}
\label{sec:main_results}

Table~\ref{tab:main_results_threshold_01} summarizes timeline reconstruction performance for text-only reconstruction and multimodal reconstruction with retrieved structured EHR evidence. Across models, multimodal calibration has little effect on event match rate, indicating that structured rows do not primarily expand the set of events recovered from narrative text. Instead, structured evidence primarily affects temporal localization, with effects that vary by model and metric. These results support the role of structured rows as temporal evidence for text-derived events, rather than as a substitute event source.

\rebuttal{
Mean paired case-level temporal differences generally favored multimodal calibration: 4 of 6 models showed higher concordance and 3 of 6 showed higher AULTC, while no model showed lower performance on either temporal metric. However, uncertainty was substantial: no paired confidence interval for the multimodal--unimodal concordance difference excluded zero, and only DeepSeek V3.2 showed an AULTC interval entirely above zero ((\hyperref[supp:paired-bootstrap]{Supplementary Material \S\ref*{supp:paired-bootstrap}}, \hyperref[tab:supp-paired-bootstrap]{Table~\ref*{tab:supp-paired-bootstrap}})). Performance was generally higher on the i2b2-derived subset, but modality effects were source-dependent ((\hyperref[supp:dataset-stratified]{Supplementary Material \S\ref*{supp:dataset-stratified}}, Tables~\hyperref[tab:supp-i2b2]{\ref*{tab:supp-i2b2}}--\hyperref[tab:supp-mimiciv]{\ref*{tab:supp-mimiciv}})).}

\begin{figure}[t]
    \centering
    \includegraphics[width=0.479\textwidth]{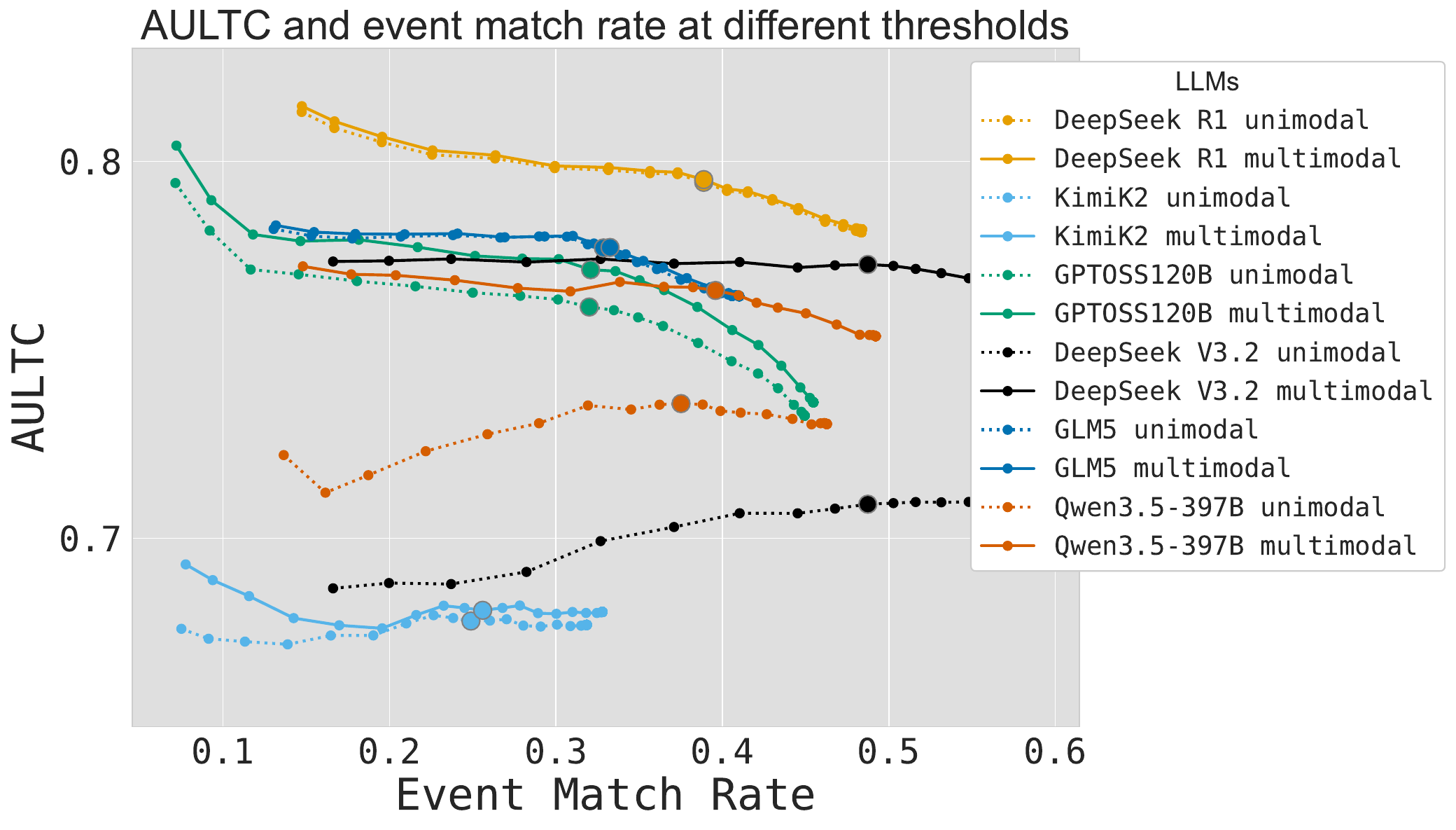}
    \hfill
    \includegraphics[width=0.479\textwidth]{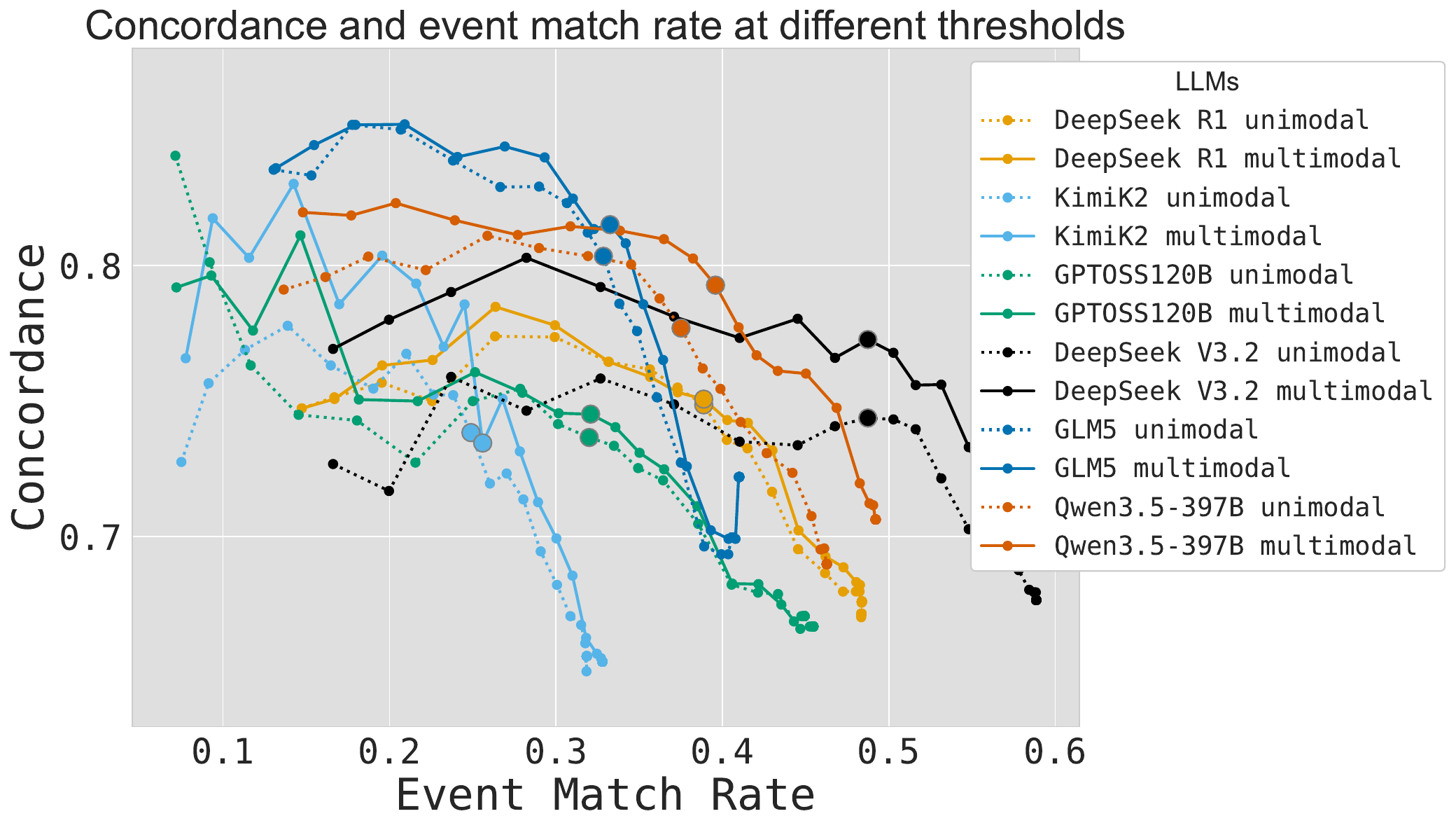}

    \vspace{-4pt}
    \begin{minipage}[t]{0.485\textwidth}
        \centering
    \end{minipage}
    \hfill
    \begin{minipage}[t]{0.485\textwidth}
        \centering
    \end{minipage}

    \vspace{-8mm}
    \caption{Temporal performance versus event match rate across event-matching thresholds for unimodal and multimodal variants of each model on the 40-summary evaluation set. (\textbf{Left}) AULTC versus event match rate. (\textbf{Right}) Temporal concordance versus event match rate.}
    \label{fig:threshold_sweeps_main}
    \vspace{-3.3mm}
\end{figure}

Because event match rate depends on the semantic matching threshold used to align predicted and reference events, we also vary the event-matching threshold and examine the relationship between event recovery and temporal quality. Figure~\ref{fig:threshold_sweeps_main} shows AULTC and concordance as functions of event match rate across thresholds. Across operating points, multimodal reconstruction generally occupies a similar event-recovery range as text-only reconstruction while changing temporal performance, especially AULTC, for several models. This supports the same broad conclusion as Table~\ref{tab:main_results_threshold_01}: retrieved structured rows primarily affect temporal calibration rather than event discovery.

\begin{table}[t]
{\footnotesize
\setlength{\tabcolsep}{3pt}
\renewcommand{\arraystretch}{1.12}
\tbl{Performance on 40 discharge summaries at event-matching threshold $0.1$ for unimodal and multimodal timeline reconstruction. Values are point estimates with bootstrap 95\% confidence intervals in brackets. \textbf{Bold} indicates the best point estimate within each metric--modality column.}
{
\begin{adjustbox}{max width=\textwidth}
\begin{tabular}{lcccccc}
\hline
\multirow{2}{*}{\textbf{Model}} 
& \multicolumn{2}{c}{\textbf{Event match rate}} 
& \multicolumn{2}{c}{\textbf{Concordance}} 
& \multicolumn{2}{c}{\textbf{AULTC}} \\
\cline{2-7}
& \textbf{Unimodal} & \textbf{Multimodal} 
& \textbf{Unimodal} & \textbf{Multimodal} 
& \textbf{Unimodal} & \textbf{Multimodal} \\
\hline
DeepSeek R1
& 0.39 [0.32, 0.47]
& 0.39 [0.32, 0.47]
& 0.75 [0.69, 0.79]
& 0.75 [0.70, 0.80]
& \textbf{0.80 [0.76, 0.82]} 
& \textbf{0.79 [0.77, 0.82]}\\

KimiK2-Instruct
& 0.26 [0.21, 0.29]
& 0.25 [0.22, 0.30]
& 0.73 [0.69, 0.76]
& 0.74 [0.68, 0.81]
& 0.68 [0.61, 0.73]
& 0.68 [0.62, 0.73] \\

GPT-OSS-120B
& 0.32 [0.27, 0.37]
& 0.32 [0.28, 0.37]
& 0.74 [0.70, 0.77]
& 0.75 [0.71, 0.80]
& 0.76 [0.72, 0.80] 
& 0.77 [0.74, 0.81]\\

DeepSeek V3.2
& \textbf{0.49 [0.43, 0.56]}
& \textbf{0.49 [0.43, 0.56]}
& 0.74 [0.70, 0.80]
& 0.77 [0.73, 0.81]
& 0.71 [0.64, 0.78] 
& 0.77 [0.73, 0.81]\\

GLM5
& 0.33 [0.27, 0.39]
& 0.33 [0.27, 0.40]
& \textbf{0.80 [0.78, 0.83]}
& \textbf{0.82 [0.75, 0.84]}
& 0.78 [0.73, 0.82] 
& 0.78 [0.73, 0.82]\\

Qwen3.5-397B-A17B
& 0.38 [0.28, 0.47]
& 0.40 [0.29, 0.51]
& 0.78 [0.72, 0.81]
& 0.79 [0.78, 0.82]
& 0.74 [0.68, 0.77] 
& 0.77 [0.74, 0.78]\\

\hline
\end{tabular}
\end{adjustbox}
}
\label{tab:main_results_threshold_01}
}
\vspace{-3mm}
\end{table}

\vspace{5pt}
\noindent\textbf{Scaffold ablations.}
Table~\ref{tab:ablation_and_DTW}A evaluates whether the staged scaffolded workflow is necessary, or whether similar performance can be obtained by generating a complete timeline in one step with retrieved structured rows.  The single-step formulation performs worse than the full scaffolded variant, suggesting that clinical timeline reconstruction benefits from first identifying temporally informative anchor events before reconstructing the full event set. The stage at which structured evidence is introduced also matters. Final-stage calibration yields the highest event match rate, but the strongest temporal metrics are obtained when structured evidence is used both to calibrate the anchor scaffold and to refine the final assembled timeline. This supports the design choice of using structured rows first to stabilize the temporal backbone of the encounter and then to recalibrate the expanded trajectory.

\begin{table}[h] 
\setlength{\tabcolsep}{3pt} \renewcommand{\arraystretch}{1.08} 
\tbl{\rebuttal{\textbf{(A)} DeepSeek V3.2 scaffold and calibration-stage ablations with bootstrap 95\% confidence intervals. \textbf{(B)} Retrieval calibration compared with sentence-embedding and lexical Dynamic Time Warping (DTW) alignments.}}
{ 
\begin{minipage}{\textwidth} 
\scriptsize 
\noindent\textbf{A. Scaffold and calibration-stage ablations}\par
\vspace{2pt}
\begin{tabular*}{\textwidth}{@{\extracolsep{\fill}}lcccccc@{}} 
\hline 
\multirow{2}{*}{\textbf{Ablation}} 
& \multicolumn{2}{c}{\textbf{Event match rate}} 
& \multicolumn{2}{c}{\textbf{Concordance}} 
& \multicolumn{2}{c}{\textbf{AULTC}} \\ 
\cline{2-7} 
& \textbf{Unimodal} 
& \textbf{Multimodal} 
& \textbf{Unimodal} 
& \textbf{Multimodal} 
& \textbf{Unimodal} 
& \textbf{Multimodal} \\ 
\hline 
single-step 
& 0.39 [0.35, 0.45] 
& 0.39 [0.35, 0.44] 
& 0.70 [0.65, 0.74] 
& 0.71 [0.65, 0.75] 
& \textbf{0.78 [0.74, 0.83]} 
& \textbf{0.77 [0.73, 0.83]} \\ 
central only 
& -- 
& 0.46 [0.42, 0.51] 
& -- 
& 0.72 [0.66, 0.77] 
& -- 
& 0.75 [0.71, 0.80] \\ 
final only 
& \textbf{0.51 [0.46, 0.56]} 
& \textbf{0.52 [0.46, 0.58]} 
& 0.68 [0.64, 0.72] 
& 0.69 [0.64, 0.72] 
& 0.72 [0.67, 0.78] 
& 0.73 [0.69, 0.78] \\ 
central + final 
& 0.49 [0.43, 0.56]
& 0.49 [0.43, 0.56]
& \textbf{0.74 [0.70, 0.80]}
& \textbf{0.77 [0.73, 0.81]}
& 0.71 [0.64, 0.78] 
& \textbf{0.77 [0.73, 0.81]} \\
\hline 
\end{tabular*}

\vspace{2pt}
\noindent\textit{Note:} Central-only, final-only, and central + final correspond to updating the anchor scaffold only, the final timeline only, or both stages, respectively. The central-only condition yields a single final timeline and is undefined for the unimodal setting.
\vspace{5pt}

\noindent\textbf{B. Retrieval calibration vs. explicit temporal alignment (Dynamic Time Warping)}

\vspace{2pt}
\begin{adjustbox}{max width=\textwidth}
\begin{tabular}{@{}l*{9}{c}@{}}
\hline
\multirow{2}{*}{\textbf{Model}}
& \multicolumn{3}{c}{\textbf{Event match rate}}
& \multicolumn{3}{c}{\textbf{Concordance}}
& \multicolumn{3}{c}{\textbf{AULTC}} \\
\cline{2-4}\cline{5-7}\cline{8-10}
& \textbf{Retrieval}
& \textbf{DTW(Sent)}
& \textbf{DTW(Lex)}
& \textbf{Retrieval}
& \textbf{DTW(Sent)}
& \textbf{DTW(Lex)}
& \textbf{Retrieval}
& \textbf{DTW(Sent)}
& \textbf{DTW(Lex)} \\
\hline
DeepSeek R1
& 0.39 & 0.39 & 0.39
& 0.75 & 0.75 & 0.74
& 0.79 & 0.80 & 0.79 \\

KimiK2-Instruct
& 0.25 & 0.26 & 0.26
& 0.74 & 0.74 & 0.73
& 0.68 & 0.68 & 0.68 \\

GPT-OSS-120B
& 0.32 & 0.32 & 0.32
& 0.75 & 0.73 & 0.73
& 0.77 & 0.77 & 0.77 \\

DeepSeek V3.2
& 0.49 & 0.49 & 0.49
& 0.77 & 0.76 & 0.77
& 0.77 & 0.77 & 0.77 \\

GLM5
& 0.33 & 0.33 & 0.33
& 0.82 & 0.81 & 0.81
& 0.78 & 0.78 & 0.77 \\

Qwen3.5-397B-A17B
& 0.40 & 0.40 & 0.40
& 0.79 & 0.79 & 0.79
& 0.77 & 0.77 & 0.77 \\
\hline
\end{tabular}
\end{adjustbox}

\vspace{2pt}

\end{minipage}
}
\label{tab:ablation_and_DTW}
\vspace{-12pt}
\end{table}

\noindent\textbf{Explicit temporal alignment.}
We also compare retrieval-based calibration with dynamic time warping (DTW), an explicit temporal-alignment baseline that aligns text-derived events with structured rows as sequences (Table~\ref{tab:ablation_and_DTW}B). The two approaches achieve broadly comparable accuracy: DTW is slightly higher for some model--metric pairs, while retrieval-based calibration is slightly higher for others, and the differences are small relative to the variability reflected by the bootstrap confidence intervals in Table~\ref{tab:main_results_threshold_01}.  \rebuttal{Retrieval and DTW therefore show comparable aggregate accuracy rather than quantitative superiority. Their principal distinction is structural: retrieval presents event-specific candidates and can retain a text-derived estimate when no adequate counterpart exists, whereas DTW imposes monotone sequence alignment across modalities with incomplete and asymmetric event coverage.}

\vspace{-1em}
\subsection{\rebuttal{Complementary coverage of narrative and structured records}}
\label{sec:gap_results}
\vspace{-0.25em}

The main reconstruction results show that structured rows can improve event timing when used as calibration evidence. We next ask why structured records cannot simply replace narrative text and then
\rebuttal{characterize which narrative-derived events are poorly represented in structured EHR data; because this text-to-structure analysis does not assess structured-only events, it supports complementarity rather than completeness of either modality. We compare text-derived timelines against tabular records from the same encounters. }
The analysis uses the best-performing LLM-generated event timelines as the textual view, MIMIC structured data as the tabular view, and admission or the case reference time as the shared temporal anchor. For each text-derived event, we retrieve a candidate tabular counterpart using embedding-based similarity matching and evaluate it along two dimensions: semantic adequacy and temporal alignment. Semantic adequacy is assessed using retrieval-augmented scoring on a 0--1 scale, while temporal alignment is measured as the absolute difference in hours between the text-derived timestamp and the matched tabular timestamp. Based on these criteria, each event is classified as \textbf{well captured}, \textbf{complete absence}, \textbf{temporal mismatch}, \textbf{semantic distance}, or \textbf{detail gap}. Table~\ref{tab:gap_analysis} summarizes the distribution of these categories.

\begin{table}[!t]
{
\setlength{\tabcolsep}{5pt}
\renewcommand{\arraystretch}{1.12}
\tbl{Structured-data gap analysis. Each text-derived event is compared with structured EHR counterparts from the same encounter.}
{
\begin{adjustbox}{max width=0.9\textwidth}
\scriptsize
\begin{tabular}
{p{0.15\textwidth}p{0.65\textwidth}p{0.16\textwidth}}
\hline
\textbf{Category} & \textbf{Definition} & \textbf{Count (\%)} \\
\hline
Well captured 
& Semantically adequate structured counterpart within the temporal window 
& 2,827 (33.5\%) \\

Complete absence 
& No structured counterpart identified 
& 2,963 (35.1\%) \\

Temporal mismatch 
& Semantically related structured row exists but occurs outside the temporal window 
& 1,463 (17.3\%) \\

Semantic distance 
& Retrieved row is not clinically equivalent to the narrative event 
& 801 (9.5\%) \\

Detail gap 
& Structured row omits clinically important attributes such as severity, laterality, duration, progression, or functional status 
& 385 (4.6\%) \\
\hline
Total textual events & -- & 8,439 \\
\hline
\end{tabular}
\end{adjustbox}
}
\label{tab:gap_analysis}
}
\vspace{-1em}
\end{table}

Across 8,439 text-derived events, only 33.5\% are well captured by structured data, while 35.1\% have no structured counterpart at all. When a structured counterpart exists, its timestamp is often useful: the median temporal discrepancy is 4.3 hours, with 58.4\% of matched events within 6 hours, 69.5\% within 12 hours, and 79.3\% within 24 hours. The missing events are not merely low-value narrative detail; they include symptoms, functional status, temporal qualifiers, patient-reported information, severity descriptions, and clinical context. Representative complete-absence examples include \textbf{symptom descriptions} such as ``substernal chest pain for 1 day'' and ``developed chest pain''; \textbf{denial statements} such as ``no seizure-like activity'' and ``no deformity''; \textbf{temporal patterns} such as ``pain resolved'' and ``improving clots on LE''; \textbf{severity descriptors} such as ``EF 55\%'' and ``non-small cell lung cancer stage IIIA''; and \textbf{causal or procedural context} such as ``intubated for several days after second surgery'' and ``repair of leak from pouch.'' These findings explain why structured rows are useful for temporal calibration but incomplete as representations of patient trajectories.

\vspace{-1.em}
\subsection{Qualitative case study: text supplies phenotype, tables calibrate time}
\label{sec:vignette_results}
\vspace{-0.5em}

Figure~\ref{fig:motivation_multimodal_alignment} illustrates the qualitative behavior of the pipeline using a synthetic stroke-alert case. The narrative supplies clinically decisive phenotype details, including lobar hemorrhage location, hematoma volume, aphasia, plegia, command following, absence of intraventricular hemorrhage or hydrocephalus, and repeat-CT stability. The structured rows provide sparse but precise temporal anchors, including admission time and IV andexanet alfa administration. The text-only scaffold recovers the reversal event but places it shortly after the initial CT at approximately \(+1.5\)h; retrieval-augmented calibration links the event to the medication-administration row and shifts reversal to \(+6.5\)h relative to admission.

\begin{tcolorbox}[
    breakable,
    colback=white,
    colframe=black!45,
    title=\textbf{Pipeline output: calibrated event timeline},
    fonttitle={\fontsize{8pt}{11pt}\selectfont\bfseries}
]
\scriptsize
\setlength{\parindent}{0pt}

\newcommand{\vignetteRow}[3]{%
\noindent
\begin{minipage}[t]{0.47\linewidth}\raggedright #1\end{minipage}\hfill
\begin{minipage}[t]{0.10\linewidth}\raggedleft #2\end{minipage}\hfill
\begin{minipage}[t]{0.37\linewidth}\raggedright #3\end{minipage}\par\vspace{1.5pt}
}

\vspace{-6pt}

\noindent
\begin{minipage}[t]{0.47\linewidth}\raggedright \textbf{Event}\end{minipage}\hfill
\begin{minipage}[t]{0.10\linewidth}\raggedleft \textbf{Time (h)}\end{minipage}\hfill
\begin{minipage}[t]{0.37\linewidth}\raggedright \textbf{Source / calibration note}\end{minipage}

\vspace{2pt}
\hrule
\vspace{3pt}

\vignetteRow{Last known well at 0730}{-5.5}{Text clock time relative to admission}
\vignetteRow{Found down at home}{-1.0}{Text-derived estimate}
\vignetteRow{67 years old}{0.0}{Text, admission context}
\vignetteRow{Atrial fibrillation}{0.0}{Text, admission context}
\vignetteRow{Admitted/brought in as stroke alert}{0.0}{\textbf{Table}, anchored by admission row at 1300}
\vignetteRow{Dense right arm plegia}{0.0}{Text, anchored to admission}
\vignetteRow{Follows commands}{0.0}{Text, anchored to admission}
\vignetteRow{Initial CT showed intracerebral hemorrhage}{0.5}{Text-derived estimate}
\vignetteRow{Estimated hematoma volume $\sim$55 mL}{0.5}{Text, anchored to initial CT}
\vignetteRow{Left frontal lobar location of hemorrhage}{0.5}{Text, anchored to initial CT}
\vignetteRow{Reversal agent was administered}{6.5}{\textbf{Table}, corrected by IV andexanet alfa at 1930}
\vignetteRow{Repeat CT was stable}{8.5}{Text, inferred sequentially post-reversal}

\hrule
\end{tcolorbox}

\textbf{}This case illustrates the asymmetry motivating the workflow. Text is necessary to recover clinically meaningful phenotype information, while structured rows calibrate selected time-sensitive events when relevant counterparts exist. The resulting trajectory could support human-reviewed cohort definition or eligibility logic by preserving narrative-only clinical details while grounding treatment timing in structured evidence. 
For example, the 2024 ENRICH trial comparing surgical evacuation versus medical management had eligibility criteria beyond what would be available in structured data alone\cite{pradilla2024trial}. 
Structured data could show admission indication (hemorrhagic stroke), medication events (a confirmation proxy), and functional status scores (neurologic and disability level), but the notes contain key additional ENRICH eligibility features and timing: 55 mL hematoma volume ($30<x<80$), left frontal lobar location (lobar/anterior basal ganglionic), aphasia/plegia (new onset of neurological symptoms), no IVH/hydrocephalus and stable repeat imaging (feasibility), as well as localization of the eligibility timing window (within 24 hours of last known well time).

\vspace{-0.75em}
\section{Discussion and Conclusion}
\label{sec:discussion}
\vspace{-0.5em}

Our results suggest that the primary benefit of multimodal alignment is temporal calibration rather than event discovery.
\rebuttal{
Across models, structured evidence changed event recovery little, while mean paired case-level temporal differences generally favored multimodal calibration: 7 of 12 model--metric comparisons across concordance and AULTC were positive and none were negative. However, uncertainty in the magnitude of these improvements was substantial given the 40-case sample; paired confidence intervals excluded zero only for the DeepSeek V3.2 AULTC improvement. Source-stratified estimates also differed between i2b2-derived and MIMIC-IV cases (\hyperref[supp:paired-bootstrap]{Supplementary Material \S\ref*{supp:paired-bootstrap}} and \hyperref[supp:dataset-stratified]{\S\ref*{supp:dataset-stratified}}).
Within the text-primary formulation, narrative text defines the candidate event set and structured rows provide partial temporal evidence; the results therefore establish the value of narrative content for events represented in the note, not that narrative is the primary source of all clinically meaningful events.}

The structured-data gap analysis explains why this distinction matters. Structured EHR data is not merely a noisier version of the narrative record; it offers a different view of the patient trajectory. When relevant counterparts exist, their timestamps can anchor timeline calibration. However, many clinically meaningful narrative events are absent, delayed, or semantically compressed in tabular form, including symptoms, functional status, temporal qualifiers, patient-reported information, and clinical context. 
\rebuttal{These findings characterize
asymmetric coverage within the present text-primary analysis; they do not assess structured-only events or establish that either source is complete.}

The ablation results also support the scaffolded design. One-shot timeline generation requires a model to identify clinical events, infer relative timings, resolve vague temporal expressions, and decide how to use structured evidence in a single step. By first constructing an anchor-event scaffold, the workflow separates the temporal backbone of the encounter from supporting narrative details. This mirrors clinical reasoning, where events such as admission, diagnostic testing, treatment initiation, deterioration, procedures, and discharge provide temporal reference points for interpreting the rest of the patient course. Staged calibration then allows structured evidence to stabilize the backbone before refining the full trajectory.

\vspace{5pt}
\noindent\textbf{Limitations:} 
\rebuttal{This study has several limitations. First, event match rates remain modest, so concordance and AULTC measure temporal quality conditional on successful event recovery rather than accuracy over the complete trajectory. Matched events were clinically heterogeneous, but unmatched events more often contained later-course and granular numerical documentation (\hyperref[supp:matched-unmatched]{Supplementary Material \S\ref*{supp:matched-unmatched}}, \hyperref[tab:supp-matched-unmatched]{Table~\ref*{tab:supp-matched-unmatched}}). Second, the 40-case sample provides only high-uncertainty evidence about stability across data sources and clinical
settings; moreover, each case was annotated once by one of three medically trained annotators, without duplicate annotation or inter-annotator agreement. Third, the framework emits one continuous timestamp per event; when timing is vague or absent, it infers a point estimate from context, and the model-generated certainty flag is uncalibrated (\hyperref[supp:temporal-certainty]{Supplementary Material \S\ref*{supp:temporal-certainty}}, \hyperref[tab:supp-certainty]{Table~\ref*{tab:supp-certainty}}). Fourth, the text-primary design may omit clinically meaningful structured-only events. Structured timestamps also have source-specific semantics and may represent collection, ordering, administration, result availability, device recording, or
documentation rather than clinical onset. Finally, the seven-call workflow has substantially higher inference cost than the single-step baseline.}

\vspace{5pt}
\noindent\textbf{Practical use and clinical impact:} This framework is not intended for autonomous clinical decision-making, but for temporally grounded research and clinician-reviewed workflows. Reconstructed trajectories could support retrospective cohort construction, treatment-window definition, temporal phenotyping, leakage auditing, and downstream trajectory modeling.
\rebuttal{Future work should evaluate larger, independently annotated cohorts; support interval-valued timestamps, qualitative relations, and explicit time-unknown outputs; develop bidirectional event discovery that preserves source provenance and timestamp semantics; and test stage-specific cost-quality tradeoffs. 
Overall, the findings support a retrieval-augmented view in which narrative text supplies candidate-event semantics, structured rows provide partial and source-dependent temporal evidence, and preserve-by-default calibration can improve temporal grounding for some events when relevant evidence is available.}

\vspace{-3mm}
\section*{Acknowledgements}
This research was supported by the Intramural Research Program of the National Institutes of Health (NIH) and utilized the computational resources of the \href{http://hpc.nih.gov}{NIH HPC Biowulf cluster}. The contributions of the NIH author(s) are considered Works of the United States Government. 
S.N. was supported in part by the Carnegie Mellon University TCS Presidential Fellowship, and the Natural Sciences and Engineering Research Council of Canada (NSERC) Canada Graduate Research Scholarship --- Doctoral (CGRS D) Fellowship.
S.N. was also supported in part by an appointment to the National Library of Medicine Research Program administered by the Oak Ridge Institute for Science and Education (ORISE) through an interagency agreement between the U.S. Department of Energy (DOE) and the National Library of Medicine, National Institutes of Health. ORISE is managed by ORAU under DOE contract number DE-SC0014664. 
The findings and conclusions presented in this paper are those of the authors and do not necessarily reflect the views of the NIH, NLM, DOE, ORAU/ORISE or the U.S. Department of Health and Human Services. 

\bibliographystyle{ws-procs11x85}
\bibliography{references}

\begin{thebibliography}{10}

\bibitem{kyriazopoulou2021procalcitonin}
E.~Kyriazopoulou, L.~Liaskou-Antoniou, G.~Adamis, A.~Panagaki, N.~Melachroinopoulos, E.~Drakou, K.~Marousis, G.~Chrysos, A.~Spyrou, N.~Alexiou {\em et~al.}, Procalcitonin to reduce long-term infection-associated adverse events in sepsis. a randomized trial, {\em American Journal of Respiratory and Critical Care Medicine} {\bf 203}, 202  (2021).

\bibitem{seymour2019derivation}
C.~W. Seymour, J.~N. Kennedy, S.~Wang, C.-C.~H. Chang, C.~F. Elliott, Z.~Xu, S.~Berry, G.~Clermont, G.~Cooper, H.~Gomez {\em et~al.}, Derivation, validation, and potential treatment implications of novel clinical phenotypes for sepsis, {\em JAMA} {\bf 321}, 2003  (2019).

\bibitem{henry2022factors}
K.~E. Henry, R.~Adams, C.~Parent, H.~Soleimani, A.~Sridharan, L.~Johnson, D.~N. Hager, S.~E. Cosgrove, A.~Markowski, E.~Y. Klein {\em et~al.}, Factors driving provider adoption of the trews machine learning-based early warning system and its effects on sepsis treatment timing, {\em Nature Medicine} {\bf 28}, 1447  (2022).

\bibitem{kamran2024evaluation}
F.~Kamran, D.~Tjandra, A.~Heiler, J.~Virzi, K.~Singh, J.~E. King, T.~S. Valley and J.~Wiens, Evaluation of sepsis prediction models before onset of treatment, {\em NEJM AI} {\bf 1}  (2024).

\bibitem{noroozizadeh2023temporal}
S.~Noroozizadeh, J.~C. Weiss and G.~H. Chen, Temporal supervised contrastive learning for modeling patient risk progression, {\em Proceedings of Machine Learning Research} {\bf 225}, 403  (2023).

\bibitem{johnson2016mimic}
A.~E. Johnson, T.~J. Pollard, L.~Shen, L.-w.~H. Lehman, M.~Feng, M.~Ghassemi, B.~Moody, P.~Szolovits, L.~Anthony~Celi and R.~G. Mark, {MIMIC-III}, a freely accessible critical care database, {\em Scientific data} {\bf 3}, 1  (2016).

\bibitem{johnson2023mimic}
A.~Johnson, T.~Pollard, S.~Horng, L.~A. Celi and R.~Mark, {MIMIC-IV-Note: Deidentified free-text clinical notes (version 2.2)}  (2023).

\bibitem{frattallone2024using}
G.~Frattallone-Llado, J.~Kim, C.~Cheng, D.~Salazar, S.~Edakalavan and J.~C. Weiss, Using multimodal data to improve precision of inpatient event timelines, {\em Advances in Knowledge Discovery and Data Mining} {\bf 14648}, 322  (2024).

\bibitem{noroozizadeh2026reconstructing}
S.~Noroozizadeh and J.~C. Weiss, Reconstructing sepsis trajectories from clinical case reports using llms: The textual time series corpus for sepsis, {\em Proceedings of Machine Learning Research} {\bf 333}, 10  (2026).

\bibitem{noroozizadeh2026forecasting}
S.~Noroozizadeh, S.~Kumar and J.~C. Weiss, Forecasting clinical risk from textual time series: Structuring narratives for temporal ai in healthcare, {\em Proceedings of the AAAI Conference on Artificial Intelligence} {\bf 40}, 39080  (2026).

\bibitem{moldwin2021empirical}
A.~Moldwin, D.~Demner-Fushman and T.~R. Goodwin, Empirical findings on the role of structured data, unstructured data, and their combination for automatic clinical phenotyping, {\em AMIA Joint Summits on Translational Science Proceedings} {\bf 2021}, 445  (2021).

\bibitem{seinen2025using}
T.~M. Seinen, J.~A. Kors, E.~M. van Mulligen and P.~R. Rijnbeek, Using structured codes and free-text notes to measure information complementarity in electronic health records: Feasibility and validation study, {\em Journal of Medical Internet Research} {\bf 27}, p. e66910  (2025).

\bibitem{kohane1987temporal}
I.~S. Kohane, Temporal reasoning in medical expert systems, in {\em {MEDINFO} 86: Proceedings of the Fifth World Congress on Medical Informatics\/},  eds. R.~Salamon, B.~Blum and M.~J{\o}rgensen (North-Holland, Amsterdam, 1986).

\bibitem{zhou2007temporal}
L.~Zhou and G.~Hripcsak, Temporal reasoning with medical data—a review with emphasis on medical natural language processing, {\em Journal of Biomedical Informatics} {\bf 40}, 183 (Apr 2007).

\bibitem{uzuner20112010}
{\"O}.~Uzuner, B.~R. South, S.~Shen and S.~L. DuVall, 2010 i2b2/va challenge on concepts, assertions, and relations in clinical text, {\em Journal of the American Medical Informatics Association} {\bf 18}, 552  (2011).

\bibitem{sun2013evaluating}
W.~Sun, A.~Rumshisky and O.~Uzuner, Evaluating temporal relations in clinical text: 2012 i2b2 challenge, {\em Journal of the American Medical Informatics Association} {\bf 20}, 806  (2013).

\bibitem{sun2013annotating}
W.~Sun, A.~Rumshisky and O.~Uzuner, Annotating temporal information in clinical narratives, {\em Journal of biomedical informatics} {\bf 46}, S5  (2013).

\bibitem{styler2014temporal}
I.~Styler, William~F., S.~Bethard, S.~Finan, M.~Palmer, S.~Pradhan, P.~C. de~Groen, B.~Erickson, T.~Miller, C.~Lin, G.~Savova and J.~Pustejovsky, Temporal annotation in the clinical domain, {\em Transactions of the Association for Computational Linguistics} {\bf 2}, 143  (2014).

\bibitem{lin2013medtime}
Y.-K. Lin, H.~Chen and R.~A. Brown, Medtime: A temporal information extraction system for clinical narratives, {\em Journal of biomedical informatics} {\bf 46}, S20  (2013).

\bibitem{savova2010mayo}
G.~K. Savova, J.~J. Masanz, P.~V. Ogren, J.~Zheng, S.~Sohn, K.~C. Kipper-Schuler and C.~G. Chute, Mayo clinical text analysis and knowledge extraction system ({cTAKES}): architecture, component evaluation and applications, {\em Journal of the American Medical Informatics Association} {\bf 17}, 507  (2010).

\bibitem{lin2016multilayered}
C.~Lin, D.~Dligach, T.~A. Miller, S.~Bethard and G.~K. Savova, Multilayered temporal modeling for the clinical domain, {\em Journal of the American Medical Informatics Association} {\bf 23}, 387  (2016).

\bibitem{nikfarjam2013towards}
A.~Nikfarjam, E.~Emadzadeh and G.~Gonzalez, Towards generating a patient’s timeline: extracting temporal relationships from clinical notes, {\em Journal of biomedical informatics} {\bf 46}, S40  (2013).

\bibitem{hirsch2015harvest}
J.~S. Hirsch, J.~S. Tanenbaum, S.~Lipsky~Gorman, C.~Liu, E.~Schmitz, D.~Hashorva, A.~Ervits, D.~Vawdrey, M.~Sturm and N.~Elhadad, {HARVEST}, a longitudinal patient record summarizer, {\em Journal of the American Medical Informatics Association} {\bf 22}, 263  (2015).

\bibitem{leeuwenberg2020towards}
A.~Leeuwenberg and M.-F. Moens, Towards extracting absolute event timelines from english clinical reports, {\em IEEE/ACM Transactions on Audio, Speech, and Language Processing} {\bf 28}, 2710  (2020).

\bibitem{wang2025large}
J.~Wang and J.~C. Weiss, A large-language model framework for relative timeline extraction from pubmed case reports  (2025).

\bibitem{kumar2026temporally}
S.~Kumar and J.~C. Weiss, Temporally phenotyping glp-1ra case reports with large language models: A textual time series corpus and risk modeling, {\em medRxiv} , 2026  (2027).

\bibitem{yao2025chemotimelines}
J.~Yao, H.~Hochheiser, W.~Yoon, E.~T. Goldner and G.~K. Savova, Overview of the 2025 shared task on chemotherapy treatment timeline extraction, in {\em Proceedings of the 7th Clinical Natural Language Processing Workshop\/},  eds. A.~Ben~Abacha, S.~Bethard, D.~Bitterman, T.~Naumann and K.~Roberts (Association for Computational Linguistics, Virtual, oct 2025).

\bibitem{zhang2025uwbionlp}
T.~M. Zhang, Z.~Sun, S.~Zeng, C.~Li, N.~F. Abernethy, B.~D. Lam, F.~Xia and M.~Yetisgen, {UW}-{B}io{NLP} at {C}hemo{T}imelines 2025: Thinking, fine-tuning, and dictionary-enhanced {LLM} systems for chemotherapy timeline extraction, in {\em Proceedings of the 7th Clinical Natural Language Processing Workshop\/},  eds. A.~Ben~Abacha, S.~Bethard, D.~Bitterman, T.~Naumann and K.~Roberts (Association for Computational Linguistics, Virtual, oct 2025).

\bibitem{liu2022multimodal}
S.~Liu, X.~Wang and et~al., Multimodal data matters: language model pre-training over structured and unstructured electronic health records, {\em IEEE Journal of Biomedical and Health Informatics} {\bf 27}, 504  (2022).

\bibitem{noroozizadeh2025pmoa}
S.~Noroozizadeh, S.~Kumar, G.~H. Chen and J.~C. Weiss, Pmoa-tts: Introducing the pubmed open access textual times series corpus  (2026).

\bibitem{pradilla2024trial}
G.~Pradilla, J.~J. Ratcliff, A.~J. Hall, B.~R. Saville, J.~W. Allen, G.~Paulon, A.~McGlothlin, R.~J. Lewis, M.~Fitzgerald, A.~F. Caveney {\em et~al.}, Trial of early minimally invasive removal of intracerebral hemorrhage, {\em New England Journal of Medicine} {\bf 390}, 1277  (2024).

\end{thebibliography}

\clearpage
\begingroup

\newcounter{ttasuppsection}
\renewcommand{\thettasuppsection}{S\arabic{ttasuppsection}}
\newcounter{ttasupptable}
\renewcommand{\thettasupptable}{S\arabic{ttasupptable}}

\newcommand{\TTASuppSection}[2]{%
  \par\vspace{0.8em}%
  \refstepcounter{ttasuppsection}\label{#2}%
  \noindent\textbf{\thettasuppsection. #1}\par\vspace{0.25em}%
}

\newcommand{\TTASuppTableCaption}[2]{%
  \refstepcounter{ttasupptable}\label{#1}%
  \noindent\textbf{Supplementary Table \thettasupptable.} #2\par\vspace{0.35em}%
}

\phantomsection\label{supp:material}
\begin{center}
{\LARGE\bfseries Supplementary Material}\par
\vspace{0.35em}
\end{center}

\vspace{0.5em}
This supplement provides additional methodological details and analyses. Unless otherwise stated, analyses use the event-matching threshold of 0.1 and the event definitions and temporal metrics described in the main manuscript. Higher concordance and AULTC indicate better temporal performance. Calculations use unrounded values and displayed metric values are rounded to two decimal places.

\TTASuppSection{Evaluation cohort and reference timeline construction}{supp:eval-cohort}
\noindent
The evaluation set contains 15 i2b2-derived discharge summaries and 25 MIMIC-IV discharge summaries. The i2b2-derived summaries were randomly sampled from eligible cases. The MIMIC-IV summaries were selected in two patient-level batches: for five randomly sampled patients, the first discharge summary was used; for 20 additional randomly sampled patients, one discharge summary was randomly selected. Each MIMIC-IV case therefore represents a distinct patient. Sampling was not conditioned on diagnosis, temporal richness, encounter complexity, or model performance. Eligibility required a discharge summary and structured EHR rows aligned to the same hospitalization; no additional clinical enrichment criteria were applied.

Three clinically trained annotators created the reference timelines across the 40 cases. Each case was assigned to one annotator, who reviewed the discharge summary and structured EHR record aligned to the same hospitalization, identified standalone clinical events using the event definition in the main manuscript, and assigned each event a time in hours relative to admission or the case-specific reference time. Event identity and timestamp were annotated jointly. Each case was annotated once; therefore, there was no overlapping independent annotation set, formal adjudication, or inter-annotator agreement calculation.

\TTASuppSection{Preserve-by-default calibration rule}{supp:calibration-rule}
\noindent The same retrieval and calibration logic is used for anchor calibration and final full-timeline refinement. For each narrative-derived event, the calibration model receives the event, its initial text-derived time, the source narrative, and the top-k retrieved structured rows. Each row is provided as name $\mid$ value $\mid$ time $\mid$ similarity. Calibration proceeds as follows:
\begin{enumerate}
    \item Exclude retrieved rows that are not clinically relevant to the event.
    \item If no clinically relevant row remains, retain the initial text-derived time.
    \item If a relevant merged row contains multiple timestamps, use the narrative context to select the most plausible occurrence.
    \item If relevant rows conflict, select the most clinically relevant and confident candidate.
    \item Update the event time only when the selected evidence justifies a change; otherwise, preserve the initial time.
\end{enumerate}
Cosine similarity determines which rows are retrieved and their order, but it is not an automatic update criterion. The procedure returns one final timestamp per event and does not force every narrative event to align to a structured row.

\TTASuppSection{Paired case-level bootstrap analysis}{supp:paired-bootstrap}
\vspace{2mm}
\noindent\textbf{Methods.} We compared multimodal and unimodal reconstruction using a paired case-level bootstrap with 10,000 resamples. In each replicate, the same 40 discharge-summary case IDs were sampled with replacement for both conditions. Event matching and all three evaluation metrics were recomputed within the sampled cases at threshold 0.1. For each metric, we recorded Delta = multimodal - unimodal and calculated the percentile 95\% confidence interval from the 10,000 paired differences. Resampling whole cases preserves dependence among events from the same discharge summary.

\vspace{2mm}\noindent
\textbf{Results.} The paired differences were generally small and model-dependent. No concordance confidence interval excluded zero. Event match rate decreased slightly for KimiK2-Instruct and increased slightly for Qwen3.5-397B-A17B; the unrounded intervals for these two changes excluded zero, although the displayed bounds round to 0.00. For AULTC, only DeepSeek V3.2 had a confidence interval entirely above zero. Qwen3.5-397B-A17B had a positive AULTC point estimate, but its interval included zero.

\par\medskip
\noindent\begin{minipage}{\textwidth}
\small
\vspace{0.3em}
\TTASuppTableCaption{tab:supp-paired-bootstrap}{Paired case-level bootstrap differences between multimodal and unimodal timeline reconstruction at event-matching threshold 0.1. Values are multimodal minus unimodal point estimates with paired bootstrap 95\% confidence intervals in brackets; positive values favor multimodal reconstruction.}
\centering
\setlength{\tabcolsep}{5pt}
\begin{tabular}{@{}lccc@{}}
\toprule
\textbf{Model} & \textbf{Delta Event match rate} & \textbf{Delta Concordance} & \textbf{Delta AULTC} \\
\midrule
DeepSeek R1 & 0.00 [0.00, 0.00] & 0.00 [-0.02, 0.01] & 0.00 [0.00, 0.00] \\
KimiK2-Instruct & \textbf{-0.01 [-0.01, 0.00]} & 0.00 [-0.07, 0.06] & 0.00 [-0.01, 0.01] \\
GPT-OSS-120B & 0.00 [-0.01, 0.01] & 0.01 [-0.01, 0.03] & 0.01 [0.00, 0.03] \\
DeepSeek V3.2 & 0.00 [0.00, 0.00] & 0.03 [-0.01, 0.08] & \textbf{0.06 [0.02, 0.13]} \\
GLM5 & 0.00 [0.00, 0.01] & 0.01 [-0.02, 0.03] & 0.00 [-0.01, 0.01] \\
Qwen3.5-397B-A17B & \textbf{0.02 [0.00, 0.05]} & 0.02 [0.00, 0.05] & 0.03 [0.00, 0.07] \\
\bottomrule
\end{tabular}
\par\vspace{0.3em}
\begin{minipage}{\textwidth}
\footnotesize\textit{Note:} Each interval is based on 10,000 paired case-level bootstrap resamples. Bold indicates that the unrounded 95\% confidence interval excludes zero. A displayed bound of 0.00 may be slightly above or below zero before rounding. The paired analysis is the primary inferential comparison between modalities.
\end{minipage}
\end{minipage}\par\medskip

\TTASuppSection{Characterization of matched and unmatched reference events}{supp:matched-unmatched}
\vspace{2mm}
\noindent
\textbf{Methods.} We analyzed all 8,066 manually annotated reference events in the DeepSeek V3.2 multimodal match file. A reference event was classified as matched when a predicted counterpart was present and the semantic error rate was at most 0.1; events with a missing counterpart or error rate greater than 0.1 were classified as unmatched. Description length was counted using alphanumeric tokens. Temporal position was summarized among events with an available reference time: within 24 hours denotes an absolute time no greater than 24 hours from the case reference, and at least 7 days after denotes a reference time of at least +168 hours. An event was considered numerical when its reference-event text contained at least one digit. We also reviewed matched and unmatched event descriptions qualitatively. Temporal metrics were not calculated for unmatched events because they have no aligned predicted timestamp.

\vspace{2mm}
\noindent
\textbf{Results.} Of the 8,066 reference events, 3,931 were matched and 4,135 were unmatched. Among unmatched events, 3,319 had no assigned predicted counterpart and 816 had a candidate counterpart with semantic error greater than 0.1. Both subsets had a median description length of three tokens. Matched events covered diverse clinical categories rather than only admission, discharge, or procedure anchors. Unmatched events were more often located at least seven days after the case reference time and more often contained a numerical value. Qualitative review identified recurring unmatched laboratory and imaging details, medication doses and discharge prescriptions, pertinent negatives, follow-up and functional or social details, and context-dependent fragments that were difficult to recover as standalone events.

\par\medskip
\noindent\begin{minipage}{\textwidth}
\small

\TTASuppTableCaption{tab:supp-matched-unmatched}{
Descriptive comparison of matched and unmatched reference events for DeepSeek V3.2
at event-matching threshold 0.1.
}

\centering
\begin{minipage}{0.99\linewidth}
\centering

\begin{tabular}{@{}p{0.62\linewidth}
                >{\centering\arraybackslash}p{0.17\linewidth}
                >{\centering\arraybackslash}p{0.17\linewidth}@{}}
\toprule
\textbf{Characteristic} &
\textbf{Matched} &
\textbf{Unmatched} \\
\midrule
Reference events, $n$ &
3,931 &
4,135 \\

Description length, median [IQR] tokens &
3 [2, 5] &
3 [2, 6] \\

Within 24 h of the case reference time &
52.4\% &
46.5\% \\

At least 7 d after the case reference time &
19.7\% &
28.6\% \\

Contains a numerical value &
18.0\% &
36.8\% \\
\bottomrule
\end{tabular}

\par\vspace{0.3em}

\footnotesize
\raggedright
\textit{Note:} Among unmatched events, 3,319 (80.3\%) had no assigned predicted
counterpart and 816 (19.7\%) had a candidate counterpart with semantic error greater
than 0.1. Temporal-position percentages are calculated among events with an available
reference timestamp. This is a descriptive coverage analysis; timestamp error,
concordance, and AULTC are undefined for unmatched events.

\end{minipage}
\end{minipage}
\par\medskip

\TTASuppSection{Dataset-stratified performance}{supp:dataset-stratified}
\vspace{2mm}\noindent
\textbf{Methods.} We repeated the primary evaluation separately for the 15 i2b2-derived cases and 25 MIMIC-IV cases using event-matching threshold 0.1. Event match rate, concordance, and AULTC were recomputed within each source and reported separately for unimodal and multimodal reconstruction. Source-specific 95\% confidence intervals were obtained by resampling cases with replacement within each source using the same 200-resample case-level bootstrap procedure as the marginal confidence intervals in the main analysis. These tables provide descriptive source-specific estimates; they are not paired tests of the multimodal-minus-unimodal difference.

\vspace{2mm}\noindent
\textbf{Results.} Multimodal calibration changed event match rate little within either source. Performance was generally higher on the i2b2-derived subset: five of six models had higher event match rates and all six had higher AULTC than on MIMIC-IV. Temporal shifts were heterogeneous. On i2b2, multimodal calibration produced mostly small AULTC gains and modest concordance changes. On MIMIC-IV, multimodal temporal point estimates were generally unchanged or lower in the supplied source-stratified outputs, with the largest decreases for DeepSeek V3.2 and Qwen3.5-397B-A17B. The broad overlap of the marginal intervals warrants cautious interpretation.

\par\medskip
\noindent\begin{minipage}{\textwidth}
\TTASuppTableCaption{tab:supp-i2b2}{Performance on the i2b2-derived subset (n = 15) at event-matching threshold 0.1 for unimodal and multimodal timeline reconstruction.}
\centering
\scriptsize
\begin{adjustbox}{max width=\textwidth}
\begin{tabular}{@{}lcccccc@{}}
\toprule
& \multicolumn{2}{c}{\textbf{Event match rate}} & \multicolumn{2}{c}{\textbf{Concordance}} & \multicolumn{2}{c}{\textbf{AULTC}} \\
\cmidrule(lr){2-3}\cmidrule(lr){4-5}\cmidrule(l){6-7}
\textbf{Model} & \textbf{Unimodal} & \textbf{Multimodal} & \textbf{Unimodal} & \textbf{Multimodal} & \textbf{Unimodal} & \textbf{Multimodal} \\
\midrule
DeepSeek R1 & 0.51 [0.42, 0.61] & 0.51 [0.42, 0.61] & 0.77 [0.71, 0.84] & 0.77 [0.70, 0.83] & 0.82 [0.80, 0.85] & 0.82 [0.80, 0.85] \\
KimiK2-Instruct & 0.40 [0.33, 0.50] & 0.40 [0.32, 0.50] & 0.73 [0.69, 0.81] & 0.74 [0.70, 0.80] & 0.78 [0.75, 0.80] & 0.78 [0.75, 0.81] \\
GPT-OSS-120B & 0.53 [0.48, 0.60] & 0.53 [0.48, 0.60] & \textbf{0.80 [0.76, 0.84]} & 0.77 [0.75, 0.83] & 0.80 [0.77, 0.83] & 0.81 [0.79, 0.84] \\
DeepSeek V3.2 & \textbf{0.62 [0.55, 0.70]} & \textbf{0.62 [0.55, 0.70]} & 0.77 [0.73, 0.82] & 0.79 [0.70, 0.82] & 0.82 [0.79, 0.84] & 0.82 [0.79, 0.85] \\
GLM5 & 0.28 [0.21, 0.38] & 0.28 [0.21, 0.38] & 0.80 [0.72, 0.88] & \textbf{0.81 [0.76, 0.85]} & \textbf{0.83 [0.79, 0.86]} & \textbf{0.84 [0.80, 0.87]} \\
Qwen3.5-397B-A17B & 0.61 [0.54, 0.68] & 0.61 [0.54, 0.68] & 0.78 [0.75, 0.84] & 0.78 [0.72, 0.85] & 0.80 [0.78, 0.83] & 0.81 [0.79, 0.84] \\
\bottomrule
\end{tabular}
\end{adjustbox}
\par\vspace{0.3em}
\begin{minipage}{\textwidth}
\footnotesize\textit{Note:} Values are point estimates with source-specific case-level bootstrap 95\% confidence intervals in brackets. Bold indicates the largest unrounded point estimate within each metric-modality column. All displayed values are rounded to two decimals.
\end{minipage}
\end{minipage}\par\medskip

\par\medskip
\noindent\begin{minipage}{\textwidth}
\TTASuppTableCaption{tab:supp-mimiciv}{Performance on the MIMIC-IV subset (n = 25) at event-matching threshold 0.1 for unimodal and multimodal timeline reconstruction.}
\centering
\scriptsize
\begin{adjustbox}{max width=\textwidth}
\begin{tabular}{@{}lcccccc@{}}
\toprule
& \multicolumn{2}{c}{\textbf{Event match rate}} & \multicolumn{2}{c}{\textbf{Concordance}} & \multicolumn{2}{c}{\textbf{AULTC}} \\
\cmidrule(lr){2-3}\cmidrule(lr){4-5}\cmidrule(l){6-7}
\textbf{Model} & \textbf{Unimodal} & \textbf{Multimodal} & \textbf{Unimodal} & \textbf{Multimodal} & \textbf{Unimodal} & \textbf{Multimodal} \\
\midrule
DeepSeek R1 & 0.36 [0.28, 0.45] & 0.36 [0.28, 0.45] & 0.70 [0.68, 0.80] & 0.70 [0.66, 0.80] & \textbf{0.79 [0.75, 0.82]} & \textbf{0.78 [0.75, 0.82]} \\
KimiK2-Instruct & 0.22 [0.19, 0.26] & 0.21 [0.19, 0.25] & 0.71 [0.59, 0.82] & 0.67 [0.56, 0.82] & 0.64 [0.56, 0.71] & 0.63 [0.56, 0.70] \\
GPT-OSS-120B & 0.27 [0.24, 0.34] & 0.27 [0.23, 0.34] & 0.70 [0.64, 0.75] & 0.70 [0.65, 0.77] & 0.76 [0.72, 0.80] & 0.74 [0.69, 0.79] \\
DeepSeek V3.2 & \textbf{0.46 [0.38, 0.55]} & \textbf{0.46 [0.38, 0.55]} & 0.77 [0.70, 0.83] & 0.71 [0.62, 0.77] & 0.76 [0.71, 0.81] & 0.67 [0.58, 0.76] \\
GLM5 & 0.34 [0.26, 0.41] & 0.33 [0.26, 0.40] & \textbf{0.82 [0.77, 0.85]} & \textbf{0.80 [0.73, 0.83]} & 0.77 [0.71, 0.83] & 0.76 [0.71, 0.82] \\
Qwen3.5-397B-A17B & 0.35 [0.24, 0.48] & 0.32 [0.22, 0.45] & 0.81 [0.78, 0.83] & 0.78 [0.67, 0.81] & 0.75 [0.72, 0.77] & 0.70 [0.65, 0.75] \\
\bottomrule
\end{tabular}
\end{adjustbox}
\par\vspace{0.3em}
\begin{minipage}{\textwidth}
\footnotesize\textit{Note:} Values are point estimates with source-specific case-level bootstrap 95\% confidence intervals in brackets. Bold indicates the largest unrounded point estimate within each metric-modality column. All displayed values are rounded to two decimals.
\end{minipage}
\end{minipage}\par\medskip

\TTASuppSection{Sensitivity analysis by model-generated temporal certainty}{supp:temporal-certainty}
\vspace{2mm}\noindent
\textbf{Methods.} We performed a post hoc DeepSeek V3.2 analysis using the event-level is\_certain flag derived from the model-generated confidence scores returned with relative-time estimates in the temporal-relation and event-attachment stages. We set is\_certain = 1 for confidence scores from 6 through 9 and is\_certain = 0 otherwise. Within each predicted-event stratum, we calculated the proportion of predicted events that matched a reference event and calculated AULTC over the matched events in that stratum. The matched proportion therefore has a different denominator from the reference-event match rate in the main manuscript. We did not report stratum-specific concordance because filtering to separate certainty strata produces different event-pair populations and does not support a direct comparison.

\vspace{2mm}\noindent
\textbf{Results.} The within-stratum matched proportion was 0.67 for model-labeled uncertain events and 0.64 for model-labeled certain events. Among matched events, AULTC was 0.76 for the uncertain stratum and 0.84 for the certain stratum. Thus, the certainty flag was associated with better absolute timestamp accuracy conditional on successful matching, but not with a higher probability of matching. The flag is an uncalibrated operational signal and should not be treated as a complete measure of temporal uncertainty.

\par\medskip
\noindent\begin{minipage}{\textwidth}
\small

\TTASuppTableCaption{tab:supp-certainty}{
DeepSeek V3.2 sensitivity analysis stratified by the model-generated is\_certain flag.
}
\centering
\begin{minipage}{0.99\linewidth}
\centering

\begin{tabular}{@{}p{0.42\linewidth}
                >{\centering\arraybackslash}p{0.36\linewidth}
                >{\centering\arraybackslash}p{0.14\linewidth}@{}}
\toprule
\textbf{is\_certain stratum} &
\textbf{Within-stratum matched proportion} &
\textbf{AULTC} \\
\midrule
0 (model-labeled uncertain) & 0.67 & 0.76 \\
1 (model-labeled certain)   & 0.64 & 0.84 \\
\bottomrule
\end{tabular}

\par\vspace{0.3em}

\footnotesize
\raggedright
\textit{Note:} is\_certain = 1 denotes model-assigned confidence scores from 6 through 9;
is\_certain = 0 denotes all other scores. AULTC is calculated over matched events within
each stratum. Concordance is omitted because the filtered strata yield different
event-pair populations.

\end{minipage}
\end{minipage}
\par\medskip

\TTASuppSection{Paired matched-mention sensitivity analysis}{supp:matched-mention}
\vspace{2mm}\noindent
\textbf{Methods.} This secondary analysis paired the same manually annotated reference mention across unimodal and multimodal reconstruction. Event recovery was evaluated over all 8,066 reference mentions with exact two-sided McNemar tests. For temporal accuracy, we retained only reference events matched under both conditions, compared paired absolute timestamp errors with two-sided Wilcoxon signed-rank tests, and recomputed AULTC on the same shared matched subset. Wilcoxon calculations used tie correction and dropped zero differences. No multiplicity adjustment was applied. Because mentions are clustered within 40 discharge summaries and event-rich cases contribute more observations, the reported p values are nominal and may be anti-conservative. These analyses are exploratory and secondary to the paired case-level bootstrap in \hyperref[tab:supp-paired-bootstrap]{Supplementary Table~\ref*{tab:supp-paired-bootstrap}}.

\vspace{2mm}\noindent
\textbf{Results.} Event-recovery differences were small: DeepSeek R1 and DeepSeek V3.2 were unchanged, GPT-OSS-120B changed negligibly, GLM5 and Qwen3.5-397B-A17B increased slightly, and KimiK2-Instruct decreased slightly. On the shared matched subset, AULTC point estimates favored multimodal calibration for DeepSeek V3.2, GPT-OSS-120B, GLM5, and Qwen3.5-397B-A17B; DeepSeek R1 changed negligibly and KimiK2-Instruct decreased. The nominal mention-level p values should not be interpreted as confirmatory evidence because they do not model within-case dependence.

\par\medskip
\noindent\begin{minipage}{\textwidth}
\small
\TTASuppTableCaption{tab:supp-mcnemar}{Paired matched-mention event-recovery sensitivity analysis at event-matching threshold 0.1. Values use the same pooled reference mentions under both conditions, with Delta = multimodal minus unimodal.}
\centering
\begin{tabular}{@{}lcccc@{}}
\toprule
\textbf{Model} & \textbf{Unimodal} & \textbf{Multimodal} & \textbf{Delta} & \textbf{Nominal McNemar p} \\
\midrule
DeepSeek R1 & 0.39 & 0.39 & 0.00 & 1.000 \\
KimiK2-Instruct & 0.26 & 0.25 & -0.01 & $<0.001$ \\
GPT-OSS-120B & 0.32 & 0.32 & 0.00 & 0.558 \\
DeepSeek V3.2 & 0.49 & 0.49 & 0.00 & 1.000 \\
GLM5 & 0.33 & 0.33 & 0.00 & $<0.001$ \\
Qwen3.5-397B-A17B & 0.38 & 0.40 & 0.02 & $<0.001$ \\
\bottomrule
\end{tabular}
\par\vspace{0.3em}
\begin{minipage}{\textwidth}
\footnotesize\textit{Note:} Exact two-sided McNemar tests compare paired match status over all 8,066 reference mentions. Values are rounded to two decimals, so a displayed Delta of 0.00 may reflect a small nonzero unrounded difference. The nominal p values do not account for clustering of mentions within discharge summaries and are reported only as an exploratory sensitivity analysis.
\end{minipage}
\end{minipage}\par\medskip

\par\medskip
\noindent\begin{minipage}{\textwidth}
\small
\TTASuppTableCaption{tab:supp-wilcoxon}{Paired temporal-accuracy sensitivity analysis on reference events matched under both unimodal and multimodal reconstruction. Delta AULTC = multimodal minus unimodal.}
\centering
\begin{adjustbox}{max width=\textwidth}
\begin{tabular}{@{}lccccc@{}}
\toprule
\textbf{Model} & \textbf{Shared matches, n} & \textbf{AULTC (UM)} & \textbf{AULTC (MM)} & \textbf{Delta AULTC} & \textbf{Nominal Wilcoxon p} \\
\midrule
DeepSeek R1 & 3,134 & 0.80 & 0.79 & 0.00 & 0.364 \\
KimiK2-Instruct & 1,999 & 0.68 & 0.68 & -0.01 & $<0.001$ \\
GPT-OSS-120B & 2,532 & 0.76 & 0.77 & 0.01 & $<0.001$ \\
DeepSeek V3.2 & 3,930 & 0.71 & 0.77 & 0.06 & $<0.001$ \\
GLM5 & 2,650 & 0.78 & 0.78 & 0.00 & 0.003 \\
Qwen3.5-397B-A17B & 3,025 & 0.74 & 0.77 & 0.03 & $<0.001$ \\
\bottomrule
\end{tabular}
\end{adjustbox}
\par\vspace{0.3em}
\begin{minipage}{\textwidth}
\footnotesize\textit{Note:} Nominal p values are from paired two-sided Wilcoxon signed-rank tests of absolute timestamp error among events matched in both conditions; AULTC is reported on the same shared subset. The tests do not account for clustering within discharge summaries and may be anti-conservative. UM = unimodal; MM = multimodal.
\end{minipage}
\end{minipage}\par\medskip

\TTASuppSection{Implementation and reproducibility details}{supp:implementation}
\noindent
All evaluated models are open-weight models run locally through \textit{llama.cpp}; no proprietary APIs are used. The evaluated models include DeepSeek V3.2, DeepSeek R1, GLM5, Kimi K2 Instruct, GPT-OSS-120B, and Qwen3.5 397B. The full multimodal workflow requires seven LLM calls per discharge summary in the no-retry setting, compared with one call for the single-step baseline. On discharge summaries averaging approximately 15k characters, average prompt-plus-output lengths were approximately 64k--72k tokens, with end-to-end runtimes of approximately 10--18 minutes per summary depending on model and hardware. Models were run on NVIDIA A100 GPUs, with GPU count varying by model size. The stages differ in complexity: constrained extraction and assembly may be amenable to smaller models or deterministic processing, whereas relation estimation and ambiguous calibration likely require stronger models; this stage-specific cost--quality tradeoff remains to be evaluated.

\TTASuppSection{Overall interpretation of the supplementary analyses}{supp:overall-interpretation}
\noindent
Taken together, the supplementary analyses support a qualified interpretation of multimodal calibration. Structured evidence changes event recovery little and does not produce a consistently supported concordance improvement. The clearest paired case-level result is an AULTC gain for DeepSeek V3.2, indicating improved absolute timestamp placement for that model. Source-stratified estimates and the matched-mention sensitivity analysis show that effects vary by model, dataset, metric, and analysis population. Temporal metrics remain conditional on successful event recovery, and the model-generated certainty flag does not fully represent temporal uncertainty. These results therefore support describing the multimodal effects as modest and model-dependent rather than uniform.

\endgroup

\end{document}